\documentclass[twocolumn, switch]{article} 
\UseRawInputEncoding
\usepackage{preprint}

\usepackage{multirow}
\usepackage{multicol}
\usepackage{makecell}
\usepackage{csvsimple}
\usepackage{siunitx}
\usepackage{bbm}
\usepackage{bbold}
\usepackage{amsmath}
\usepackage{amssymb}

\usepackage{graphicx}
\graphicspath{ {./images/} }

\usepackage{subcaption}
\usepackage[font=small]{caption}

\usepackage{algorithm}
\usepackage{algpseudocode}

\usepackage{rotating}
\usepackage[T1]{fontenc}

\usepackage{caption} 
\usepackage[stable]{footmisc}
\usepackage{url}

\author{%
\begin{tabular}{c} Chenhao Tong \\ The University of Melbourne \\ Melbourne, VIC \\ chetong@unimelb.edu.au \end{tabular} \and
\and
\begin{tabular}{c} Maria A. Rodriguez \\ The University of Melbourne \\ Melbourne, VIC \\ marodriguez@unimelb.edu.au \end{tabular} 
\and
\begin{tabular}{c} Richard O. Sinnott \\ The University of Melbourne \\ Melbourne, VIC  \\ rsinnott@unimelb.edu.au\\
\end{tabular}}

\title{Autonomous Vehicle Patrolling Through Deep Reinforcement Learning: Learning to Communicate and Cooperate}

\begin{document}

\twocolumn[ 
    \begin{@twocolumnfalse} 
        \maketitle
        \begin{abstract}
            Autonomous vehicles are suited for continuous area patrolling problems. Finding an optimal patrolling strategy can be challenging due to unknown environmental factors, such as wind or landscape; or autonomous vehicles' constraints, such as limited battery life or hardware failures. Importantly, patrolling large areas often requires multiple agents to collectively coordinate their actions. However, an optimal coordination strategy is often non-trivial to be manually defined due to the complex nature of patrolling environments. In this paper, we consider a patrolling problem with environmental factors, agent limitations, and three typical cooperation problems -- collision avoidance, congestion avoidance, and patrolling target negotiation. We propose a multi-agent reinforcement learning solution based on a reinforced inter-agent learning (RIAL) method. With this approach, agents are trained to develop their own communication protocol to cooperate during patrolling where faults can and do occur. The solution is validated through simulation experiments and is compared with several state-of-the-art patrolling solutions from different perspectives, including the overall patrol performance, the collision avoidance performance, the efficiency of battery recharging strategies, and the overall fault tolerance.
        \end{abstract}
    \end{@twocolumnfalse}
]


\section{Introduction} \label{sec:intro}
\emph{Multi-agent Patrolling} (MAP) can be defined as a group of agents travelling regularly through an area so that emerging events of interest can be identified as early as possible. Solutions to the patrolling problem aim to minimize the time between visits to any location in a given area \cite{p1-first}. Finding an optimal patrolling solution is non-trivial. Chevaleyre~\cite{p2-theoretical} demonstrates that the patrolling problem is highly related to the well-known Travelling Salesman Problem (TSP) and thus is NP-hard. In addition, real patrolling environments contain non-deterministic factors, such as wind or landscape, which will affect the patrolling vehicles' movements. Further, patrolling vehicles can have constraints, such as hardware failures or energy constraints. Importantly, patrolling large areas often requires multiple vehicles to coordinate their actions collectively.

Existing solutions can be largely divided into pre-defined route strategies and dynamic cooperation strategies \cite{recent_trend}. In pre-defined route strategies, agents calculate their patrolling route prior to patrolling and follow the route during runtime. However, these solutions are not ideal for dynamic patrolling environments, as environmental factors may well alter agents' movements, and thus preventing them from following the pre-calculated patrol routes. To address this, dynamic patrolling strategies have been proposed, where agents share information with each other and follow pre-defined coordination strategies to negotiate their next patrolling targets at runtime.

\begin{figure}[!hbt]

    \begin{subfigure}[t]{.45\linewidth}
        \centering
        \includegraphics[width=.45\linewidth]{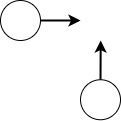}
        \caption{Example of collision scenarios}
        \label{fig:col_a}
    \end{subfigure}\hfill
    \begin{subfigure}[t]{.45\linewidth}
        \centering
        \includegraphics[width=.45\linewidth]{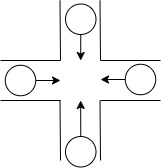}
        \caption{Example of congestion (deadlock) scenarios}
        \label{fig:con_a}
    \end{subfigure}
    \caption{Examples of multi-agent cooperation problems. The circle represents the agent, and the arrow represents the agents desired movement.}
\end{figure}

However, patrolling in real-world environments often involves cooperation challenges. Examples of typical cooperation problems considered in the literature \cite{partition-algorithm, p4-auction-review2004, rl_recent_2} include: 1) collision avoidance (Fig.~\ref{fig:col_a}) -- where vehicles need to negotiate and alter their route to avoid colliding into each other 2) congestion avoidance (Fig.~\ref{fig:con_a}) -- where vehicles need to negotiate with one another to avoid blocking each other and causing congestion, and 3) patrolling targets negotiation -- where vehicles need to negotiate with each other regarding different patrolling targets to maximise the patrolling efficiency. Due to the high complexity of cooperation scenarios, it is non-trivial to manually define vehicles' coordination strategies, as both the optimal communication protocol and the information that needs to be shared are non-trivial to discover. Many existing solutions only take into account one aspect of the cooperation problem and, therefore, have a sub-optimal performance in real patrolling environments.

On the other hand, Multi-Agent Reinforcement Learning (MARL) is a suitable technique for solving MAP problems with complex cooperation scenarios. It enables agents to develop cooperation and patrolling strategies through trial and error with associated rewards. In this paper, we propose a MARL-based solution incorporating reinforced inter-agent learning (RIAL) method \cite{RIAL} that trains agents to develop their own communication protocol in order to patrol while considering typical cooperation problems and where agents can fail.  A modified multi-agent Proximal Policy Optimisation algorithm is proposed, together with a curriculum training strategy to train agents. The performance of the model, with a focus on collision avoidance performance, patrolling performance, battery charging performance, and fault tolerance, is validated through simulation and compared with several state-of-the-art solutions.


\section{Related Work} \label{sec:rw}
Machado et al.~\cite{p1-first} formalize the definition of the MAP problem as multiple agents continuously traversing a graph $G(V,E)$ with the aim of minimizing the idleness of every vertex. Here, the term \emph{idleness} refers to the time between two visits to the same vertex by an agent. Existing solutions to MAP can be classified into predefined route strategies and dynamic cooperation strategies. 

Examples of predefined route strategies include graph partitioning-based strategies \cite{p2-theoretical, partition1, partition2, partition3} and shortest cyclic-based strategies \cite{p2-theoretical, ACO1, ACO2, ACO3}. In graph partitioning-based strategies, non-overlapping partitions are pre-calculated, and each agent patrols a partition. In cyclic-based strategies, the shortest cycle connecting all vertices is pre-calculated, and agents are evenly spread on the path to patrol. However, finding the optimal partition strategy or shortest cycle is NP-hard \cite{p2-theoretical}. In addition, the predefined route strategy cannot tolerate environmental dynamics, and agents cannot follow predefined routes when their movements are affected by external environmental factors. 

On the other hand, in dynamic cooperation strategies, agents negotiate their next patrolling target at runtime. Examples include heuristic search-based strategies \cite{p1-first, p3-hs}, where a central coordinator assigns each agent a unique target; auction-based strategies \cite{p4-auction-review2004}, where agents bid for their next patrolling targets; Bayesian-based strategies \cite{dsf1, dsf2, dsf3}, where agents maximize their local gain and share their next step intention to reduce the interference between one another, and Multi-Agent Reinforcement Learning (MARL) based solutions \cite{p6-RL, rl_recent_1, rl_recent_2, rl_recent_3, p6-RL}, where agents learn their patrolling strategy through trial and error with rewards based on reinforcement learning.

However, the coordination and communication strategies of many dynamic cooperation strategies often rely on a precise model of the environment and agents. For instance, the precise distance or agents' travelling time between vertices. It is typically infeasible to model real environments accurately due to their complexity and uncertainty. In addition, it is non-trivial to pre-define optimal coordination/communication strategies in dynamic patrolling environments with various agents' constraints. For example, what protocol is required for agents to negotiate their next patrolling target, who should recharge first, or what to do when agent failures occur? It is also non-trivial to decide on the information that needs to be shared between agents for successful patrolling.

Instead of pre-defining the communication strategy for agents, in this work we adopt the reinforced inter-agent learning (RIAL) method proposed by Foerster et al.~\cite{rl_comm}. This allows agents to develop their own communication method (information encoding and communication protocol) to achieve optimal patrolling. A modified Proximal Policy Optimisation algorithm is used to train the communication strategy.


\section{Problem Modelling} \label{sec:modelling}

Following the work of Machado et al.~\cite{p1-first}, we consider the MAP problem as a set of agents $A$ continuously traversing a graph $G(V,E)$ with the goal of minimizing the idleness of every vertex. We define the following:
\begin{itemize}
    \item $Idle(v_t)$ -- the idleness of vertex $v$ at time $t$;
    \item $Idle(G_t)$ -- the idleness of graph $G(V,E)$ at time $t$, which is the average of the idleness of all vertices in the graph $G(V,E)$;
    \item $AVG^h(G)$ -- the average of $Idle(G_t)$ after $h$ steps of a patrolling scenario, and
    \item $\overline{MAX^h(G)}$ -- the highest average $Idle(v_t)$ measured at each step after $h$ steps of a given patrolling scenario.
\end{itemize}
Two commonly used optimization criteria \cite{p2-theoretical} are considered in this work: minimizing $AVG^h(G)$ and $\overline{MAX^h(G)}$, where $AVG^h(G)$ measures the average patrolling performance of the agents, and $\overline{MAX^h(G)}$ measures the worst patrolling performance of the agents.

\begin{figure}[!hbt]

    \begin{subfigure}[t]{.23\linewidth}
        \centering
        \includegraphics[width=.8\linewidth]{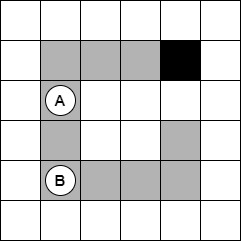}
        \caption{An example grid map}
        \label{fig:grid_map}
    \end{subfigure}\hfill
    \begin{subfigure}[t]{.23\linewidth}
        \centering
        \includegraphics[width=.8\linewidth]{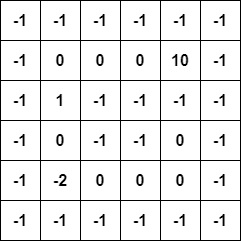}
        \caption{Matrix expression of (a) from agent $A$'s perspective}
        \label{fig:grid_map_matrix}
    \end{subfigure}
    \begin{subfigure}[t]{.23\linewidth}
        \centering
        \includegraphics[width=.8\linewidth]{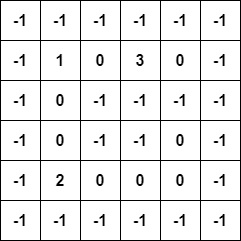}
        \caption{Priority matrix of (a)}
        \label{fig:priority_matrix}
    \end{subfigure}\hfill
    \begin{subfigure}[t]{.23\linewidth}
        \centering
        \includegraphics[width=.8\linewidth]{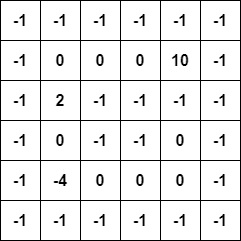}
        \caption{Message encoded in Matrix expression of (a) from agent $A$'s perspective}
        \label{fig:msg_matrix}
    \end{subfigure}

    

    \caption{An example grid map (a), its corresponding matrix expression (b) from agent A's perspective, its priority matrix (c), and its idleness matrix at timestep 0 (d).}
\end{figure}

In this work, we consider a graph that is patrolled based on a geometric map, which can be discretised into a grid. Fig.~\ref{fig:grid_map} shows an example of a grid map, where grey blocks are vertices (i.e., locations that agents can visit), white blocks are obstacles (i.e., locations that agents cannot occupy), the black block is the battery charging station, and the circles represent agents (with ID $A$ and $B$). 

The location of an agent is represented by the matrix index of the vertex, i.e. the cell occupied by the agent. This can also be the GPS coordinates. For example, in Fig.~\ref{fig:grid_map}, if the top-left block has index $(0,0)$, then the bottom-right block has index $(5,5)$, and the location of agent $A$ is $(1,2)$. The grid map can be represented as a matrix, as shown in Fig.~\ref{fig:grid_map_matrix}, where $0$ represents the vertices, $-1$ represents obstacles, and $10$ represents the battery charging station. An agent can mark its own and other agent locations on the matrix. From agent $A$'s perspective, we use $1$ to represent agent $A$'s location and $-2$ to represent agent $B$'s(other agents') location, where the negativity reflects that other agents can be considered as obstacles. It is assumed that the action space of the agent is the set $\langle \allowbreak Up,\allowbreak Down, \allowbreak Left, \allowbreak Right, \allowbreak Stay\rangle$. 

We assume that locations on the graph may have different visiting priorities, i.e., some areas may be more important to visit than others. The higher a vertex's priority, the more frequently it should be visited. The requirements for more frequent visits to high-priority vertices can be modelled through the rate of idleness-increasing of the vertices. A matrix is used to represent the vertices' priorities (Fig.\ref{fig:priority_matrix}). The priority of a vertex is represented by an integer. The priority of an obstacle is $-1$, and the priority of a battery charging station is $0$. 

Similarly, the idleness of the grid map can also be represented by a matrix. It is assumed that the idleness of vertices is $+\infty$ at the beginning of a patrolling scenario since no vertices have been visited yet. If an agent visits a vertex, the idleness of that vertex will be set to $0$. The idleness of obstacles remains constant at $-1$, and the idleness of the battery charging station remains constant at $0$. Agents can share their current locations with each other and mark the idleness of the vertices occupied by other agents to $0$ to form a global observation of the idleness matrix of the graph.

As noted, in this work, agents are trained to develop their own cooperation strategies to resolve three typical MAP cooperation problems: 1) collision avoidance, 2) congestion avoidance, and 3) patrolling target negotiation.  We use integers for agent messages. No semantic meanings are assigned to messages prior to patrolling. Through MARL, agents need to learn to 1) assign and agree on the semantic meaning of each message and 2) learn how to behave optimally when receiving messages. 

In each step, an agent's message will be broadcast to other agents. We encoded the messages of agents in a matrix expression, where an agent's message is shown in the block occupied by that agent. The sign of the message represents the ownership of a message. From an agent's perspective, the positive message is sent by itself, while the negative message is sent by others. Fig.~\ref{fig:msg_matrix} depicts an example message encoding matrix from agent $A$'s perspective, where agent $A$ sends a message $2$ and agent $B$ sends a message $4$. Agents' messages will not be carried on to the next step. Since collision avoidance is required -- agents cannot occupy the same vertex -- and therefore, no messages can overlap, and the messages of all agents can be observed. It is worth noting that the value of the battery charging station indicator should be set to be distinguished from an agent's location indicator that contains a message. For example, set as a number larger than the sum of the location indicator and the message with the highest value.

We further take into account agents' battery limitations and failure issues and assume a dynamic patrolling environment. With respect to the battery constraints of agents, agents must recharge their batteries before they run out to enable continuous patrolling. A hot-swap battery recharging scheme \cite{b-hot-swap} is assumed to recharge agents, i.e., when a vehicle goes to the battery charging station to recharge, a charged vehicle will be deployed to replace it. We do not assume that the battery hot-swap procedure can be done instantaneously. In addition, for safety purposes, modern autonomous vehicles may enforce certain behaviours if they reach low battery levels, e.g., force-landing in the case of drones\cite{dji}. Therefore, we allow to manually define the minimum battery level ($b_{l}$) whereupon the agent should stop searching and visit the battery charging station. We also assume that a battery charging station can recharge multiple agents at the same time.

Regarding agent failures, we consider catastrophic agent failure scenarios, e.g., hardware failures or running out of charge. Failed agents can no longer patrol. We assume that failed agents will be removed from the graph, or otherwise, they may block the path and cause vertices to be non-visitable, thereby affecting the ability of the remaining agents to patrol properly. The patrolling system is expected to remain functioning when one or multiple agents fail.

Regarding environmental dynamics, we consider the effects on agent movements and their battery draining rate. Specifically, the dynamics may move agents in an unintentional direction, altering the travelling time between vertices and increasing or decreasing the agent's battery consumption rate. 

In a given patrolling scenario, agents interact with the environment in the following way:

\begin{enumerate}
  \item At the beginning of a patrolling scenario, all agents are randomly placed on vertices with a random battery level remaining.
  
  \item At the beginning of a step, each agent will first make observations, then communicate, form their collective plans, and then choose an action and move to their associated corresponding vertex. After every agent arrives at their vertex, the current step is considered completed. An agent's battery will then drop a non-deterministic amount.

  \item When multiple agents head towards the same vertex, a collision occurs. Only one of the agents can occupy the vertex, and others need to return to their original position.
  
  \item At the end of each step, the idleness of all vertices will increase by the time that the agents have spent completing the step. The idleness of the vertices occupied by agents will be set to $0$.
  
  \item If the agent lands on the battery charging station intentionally, i.e., not caused by environmental dynamics, the hot-swap procedure will start immediately. Otherwise, the agent will not be replaced.

\end{enumerate}

We summarise the parameters of the environment and agents and their associated notations in Table~\ref{tab:parameters}.

\renewcommand{\arraystretch}{1.5}
\begin{table}[htb]
    \centering
    \begin{tabular}{ll} 
    \hline
    \textbf{Parameter}  & \textbf{Description} \\
    \hline
    $b_{max}$           & The maximum battery capacity of an agent. \\
    \hline
    $b^i_{init}$        & Agent $i$'s battery level at the start. \\
    \hline
    $b^i_{swap}$        & \makecell[l]{The time to replace agent $i$ with a charged  \\ agent when agent $i$ needs to recharge.} \\
    \hline
    $p^i_{dyn}$         & \makecell[l]{The non-deterministic probability that agent \\ $i$'s movement will be affected due to  \\ environmental dynamics and uncertainties.} \\
    \hline
    \end{tabular}
    \caption{The Environment and Agent Parameters.}
    \label{tab:parameters}
\end{table}


\section{Methods} \label{sec:methods}

\subsection{System Architecture}
In this work, we use the heterogeneous multi-agent system architecture proposed in \cite{my1}, where all agents execute an \emph{identical} policy based on their local observations and shared information. Since the agents are identical, the total failure scenarios of a system with $n$ agents will be $n-1$ without considering the case in which all agents fail. In contrast, in an individual learner architecture such as \cite{p6-RL, rl_recent_1}, $n$ agents result in $2^n-1$ distinct failure cases, thus requiring a vast number of training samples to cover all failure scenarios.

\subsection{Reward Function ($R$)}
A reward function reflects the performance of the agent's policy for solving patrolling tasks. In this work, the reward function evaluates agents from three perspectives: 1) the agents' patrolling performance, 2) the agents' battery charging performance, and 3) the agents' collision avoidance performance. Three reward functions, $R_p$, $R_b$, and $R_c$, evaluate the agents from each of these perspectives, as shown in Eq.~\ref{rw_highlevelreward}.

\begin{equation}
\label{rw_highlevelreward}
R = R_p + R_b + R_c
\end{equation}

\begin{figure}[!hbt]
    \centering
    \includegraphics[width=0.3\textwidth]{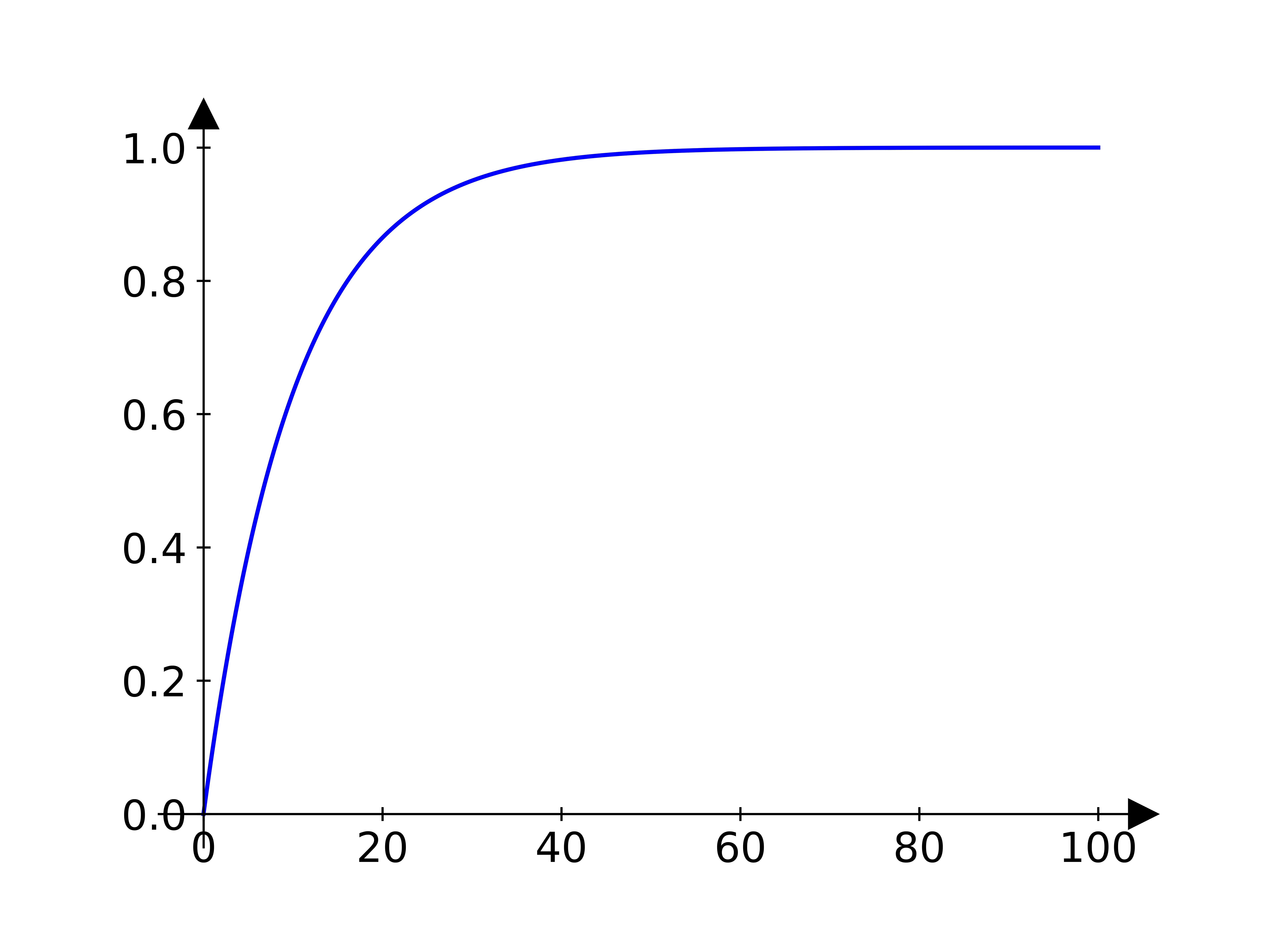}
    \caption{Function plot of $f(i) = -e^{-\frac{i}{10}} + 1$, $i \in [0,100]$, $c_{norm} = 10$}
    \label{fig:function_plot_norm}
\end{figure}

\paragraph{Patrolling performance ($R_p$)} 
During training, the idleness of vertices is normalized between $0$ and $1$ through the function $f(i) = -e^{-\frac{i}{c_{norm}}} + 1$, where $i$ is the idleness and $c_{norm}$ is a constant (Fig.~\ref{fig:function_plot_norm}). 

$R_p(k,t)$, shown in Eq.~\ref{rp}, evaluates agent $k$'s patrolling performance at time $t$. Maximising the cumulative reward of $R'_p(t)$ is equivalent to minimizing $AVG^h(G)$ and $\overline{MAX^h(G)}$, which is the agents' patrolling goal.

$D_k(t)$ is the difference reward \cite{difference_reward}, which is the difference in the rewards with ($R'_p(t)$) and without ($R'_{p,-k}(t)$) agent $k$'s action at time $t$. It reflects an agent's contribution to the patrolling problem. Therefore, the agents should contribute as much as possible to patrolling.

$c_{r'_p}$ and $c_{d}$ are scaling factors.

\begin{equation} \label{rp}
\begin{split}
&R_p(k,t) = R'_p(t) * c_{r'_p} + D_k(t) * c_{d} \\
&R'_p(t) = \frac{2-Idle(G_t)-max(Idle(v_t))}{2} \\
&D_k(t) = R'_p(t) - R'_{p,-k}(t) \\
\end{split}
\end{equation}

\paragraph{Battery usage ($R_b$)} 
An agent's battery recharging performance is evaluated based on two aspects: i) whether an agent can recharge before its battery runs out, and ii) whether an agent can recharge with a battery remaining no less than $b_l$. $R_b$ is defined as Eq.~\ref{rwb}.

\begin{equation} \label{rwb}
R_{b}(k) = 
    \begin{cases}
        -c_{pb} & \text{ $b_k$ = 0} \\
        -\frac{c_{pb_m}}{b_l}b_k + c_{pb_b} & \> 0 \leq b_k \leq b_l\\
        0 & \text{$b_k$ > $b_l$}
    \end{cases} \\
\end{equation}

If an agent runs out of battery, a penalty $c_{bp}$ will be assigned to it. Otherwise, if an agent's battery remaining $b_k$ is lower than $b_l$, a linear growing penalty, starting at value $c_{pb_b}$ with ratio $-\frac{c_{pb_m}}{b_l}$, will be assigned with respect to $b_k$.

\paragraph{Collision Avoidance ($R_c$)}
If agents collide with each other, a penalty will be assigned to all agents involved in the collision.
\begin{equation} \label{rc}
R_c = \begin{cases}
      -c_{pc} & \text{for agents involved in a collision} \\
      0 & \text{Otherwise}
    \end{cases} \\
\end{equation}

\subsection{The Learning Algorithm}
We use the Multi-agent Proximal Policy Optimization (MAPPO) algorithm \cite{ppo, multi-agent_ppo} extended from our previous work \cite{my1}. In MAPPO, agents use an actor network to approximate their policy, and a critic network to approximate the value of the state during training.

During training, we assume a maximum number of patrolling agents exist, so the size of all agents' battery information $B_t$ and location information $Loc_t$ is bounded. The observation of an agent $i$'s critic network is the set $\langle \allowbreak G(V, E), \allowbreak Idle(G_t), \allowbreak B_t, \allowbreak Loc_t\rangle$. If fewer agents than the assumed maximum are patrolling in a scenario, the vacant space in $B_t$ will be filled with $1$, and the vacant space in $Loc_t$ will be filled with the location of battery charging stations.

An agent $i$'s actor network's observation is a set $\langle \allowbreak G(V, E), \allowbreak Idle(G_t), \allowbreak b_{it}, \allowbreak Loc_t, \allowbreak ACT_{it}\rangle$. $b_{it}$ is agent $i$'s battery remaining in percentage at time $t$. $ACT_{it}$ is agent $i$'s invalid action masking set \cite{starcraft} at time $t$, which represents the validity of the agent's actions. As an example, agent $B$ in the grid map shown in Fig.~\ref{fig:grid_map} can move Up, Right, or Stay. The agent's $ACT_{Bt}$ would then be a set $<1, 0, 0, 1, 1>$. If we assume the agent's probability distribution over the action space to be $\langle 0.2, 0.1, 0.3, 0.2, 0.2 \rangle$, this distribution would then be renormalised to $\langle 0.33, 0, 0, 0.33, 0.33\rangle$ based on $ACT_{Bt}$. In real settings, the invalid action masking set can be obtained through hardware such as radar.

Regarding communication, agents are designed to have two separate actor networks, $Actor_{comm}$ and $Actor_{act}$. In each time step, $Actor_{comm}$ first chooses a message for communication. Then, based on the shared information and message, $Actor_{act}$ chooses an action to move. The RIAL method \cite{RIAL} is introduced to train agents to communicate, where the gradient of PPO is used to update parameters of the $Actor_{comm}$, although agents' actions to communication have no direct influence on agents' movement and rewards.

\begin{equation} \label{ppo_loss}
    \begin{split}
        &L^{CLIP}_{i}(\theta) = \mathbb{E}_{i,t}[min(\mathcal{R}_{i,t}(\theta) \cdot A_{i,t}, clip(\mathcal{R}_{i,t}(\theta), 1-\epsilon, 1+\epsilon)  \cdot A_{i,t})] \\
        &\text{where  } \mathcal{R}_{i,t}(\theta) = \frac{\pi_{\theta_{new}}(a_{i,t}, m_{i,t} \mid s_t)}{\pi_{\theta_{old}}(a_{i,t}, m_{i,t} \mid s_t)} \\
    \end{split}
\end{equation}

Eq.~\ref{ppo_loss} shows the loss function of the actor network in our MAPPO algorithm. $\pi_\theta(a,m|s)$ is an agent's policy that selects action $a$ and message $m$ given a state $s$ based on parameters $\theta$. $A_{i,t}$ is the General Advantage Estimate of agent $i$ at time step $t$. The $ratio_{i,t}(\theta)$ is estimated by: 

\begin{equation} \label{ppo_pi}
\frac{p_{\theta_{new}}(a_{i,t}, m_{i,t} \mid s_t)}{p_{\theta_{old}}(a_{i,t}, m_{i,t} \mid s_t)} \text{\>\>,where\>\>} p_\theta(m,a|s) =p_\theta(m|s) * p_\theta(a|s, m)
\end{equation}

$p_\theta(m|s)$ and $p_\theta(a|s, m)$ are estimated through the trajectory collected in an episode. The gradient of the loss function can, therefore, update the agent's $Actor_{comm}$ network.

\begin{figure}[!hbt]
    \begin{subfigure}[b]{.3\linewidth}
        \centering
        \includegraphics[width=.8\linewidth]{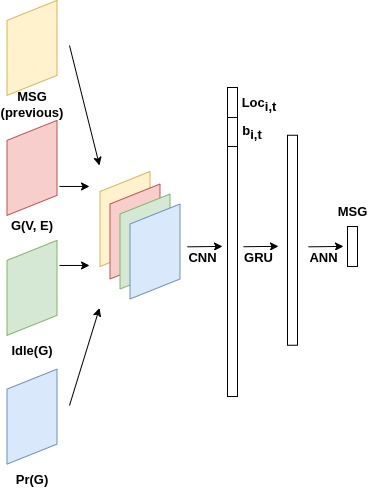}
        \caption{Actor Network (Message) - $Actor_{comm}$}
        \label{fig:msg}
    \end{subfigure}\hfill
    \begin{subfigure}[b]{.3\linewidth}
        \centering
        \includegraphics[width=.8\linewidth]{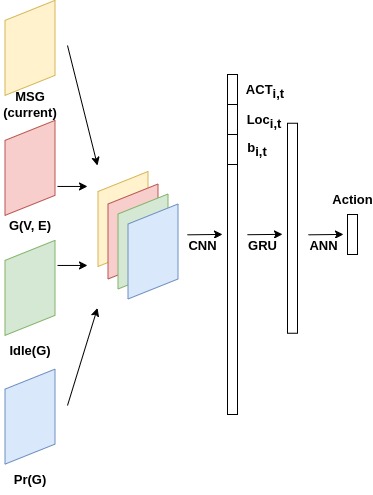}
        \caption{Actor Network (Movement) - $Actor_{act}$}
        \label{fig:actor}
    \end{subfigure}
    \begin{subfigure}[b]{.3\linewidth}
        \centering
        \includegraphics[width=.8\linewidth]{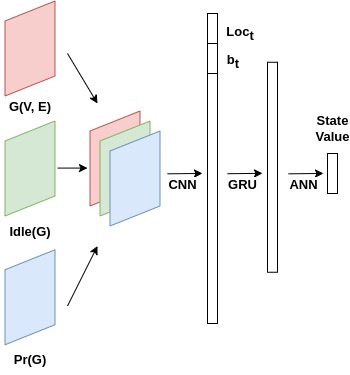}
        \caption{Critic Network}
        \label{fig:critic}
    \end{subfigure}%
    \caption{Architecture of the Actor Network ( \ref{fig:msg} \ref{fig:actor}) and the Critic Network (\ref{fig:critic}) in the MAPPO. The arrow represents the direction of the data flow.}
     \label{fig:ppo-arch}
\end{figure}

In addition, agents' negotiation on resolving a problem could last a few time steps. Therefore, a recurrent policy is used, allowing agents to memorize previous communication information. The architecture of the agents' actor and critic networks is depicted in Fig.~\ref{fig:ppo-arch}.

To train the MAPPO agents, a trajectory containing $\langle s_t, p_\theta(m_t \mid s_t), p_\theta(a_t \mid s_t, m_t), r_t, V_\theta(s_t), a_t \rangle$ is collected in each time step of a training episode. $r_t$ are agents' rewards, $V_\theta(s_t)$ are critic network's estimation of the value of state $s_t$. When an agent is recharging, we consider it to be offline and it will be replaced by a supplementary agent through battery hot-swapping. The deployed supplementary agent will then continue the trajectory collection. Since agents cannot make observations during battery hot-swapping, agents may have trajectories with different lengths. As a result, agents may calculate different values for the same state, which should be identical. As a homogeneous architecture is used such that agents have an identical policy, the missing data during the battery hot-swapping can be reconstructed based on the trajectories of other running agents.


\section{Performance evaluation}
\begin{figure}[!hbt]
    \begin{subfigure}[b]{.24\linewidth}
        \centering
        \includegraphics[width=.95\linewidth]{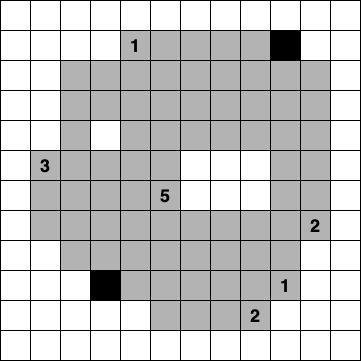}
        \caption{MAP A}
    \end{subfigure}\hfill
    \begin{subfigure}[b]{.24\linewidth}
        \centering
        \includegraphics[width=.95\linewidth]{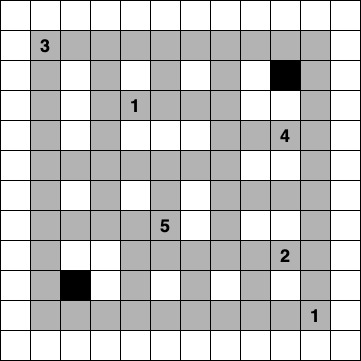}
        \caption{MAP B}
    \end{subfigure}
    \begin{subfigure}[b]{.24\linewidth}
        \centering
        \includegraphics[width=.95\linewidth]{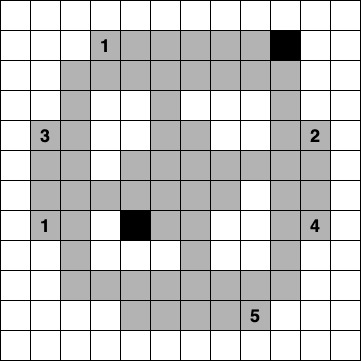}
        \caption{MAP C}
    \end{subfigure}\hfill
    \begin{subfigure}[b]{.24\linewidth}
        \centering
        \includegraphics[width=.95\linewidth]{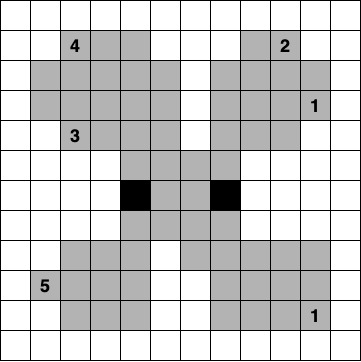}
        \caption{MAP D}
    \end{subfigure}%
    \caption{Four patrolling maps. The numbers represent the priorities of vertices, while the un-numbered vertices have priority 0 (normal priority).}
     \label{fig:maps}
\end{figure}

To provide a general performance analysis of the proposed deep MARL-based approach, we train models on $4$ patrolling maps (Fig.~\ref{fig:maps}) with $1$ to $5$ agents. We evaluate the performance of the proposed deep MARL-based strategy, including agents' ability to resolve three typical cooperation problems from the following perspectives: 

\begin{itemize}
    \item Battery recharging performance -- whether agents can successfully recharge themselves at the appropriate time,
    \item Patrolling performance -- whether agents can minimise two criteria $AVG^h(G)$ and $\overline{MAX^h(G)}$. It also reflects whether agents can properly negotiate their targets to ensure high-priority vertices are visited more frequently while low-priority vertices are also patrolled without major delays;
    \item Collision avoidance performance -- whether agents can communicate and ensure only one agent can occupy a given vertex. It also reflects agents' congestion avoidance, i.e., the ability to avoid blocking one another. The number of collisions will increase as more agents patrol.
    \item Fault tolerance -- whether the patrolling system can remain functioning when one or more agent failures occur.
\end{itemize}

Regarding the agent collision avoidance evaluation, in a real environment, a collision can cause agent hardware failures. However, in Section \ref{Sec:collision}, the simulation results show that uncoordinated strategies can cause hundreds or even thousands of collisions among agents in a 24-hour patrolling scenario. Therefore, during training and evaluation, we assume that after a collision, a randomly selected agent will occupy the vertex, and the other agent(s) will return to their original position. We test the patrolling system's fault tolerance in separate scenarios.


We compare the performance of our model (RL-MSG) with 5 patrolling strategies: 1) Conscientious Reactive strategy (CR) \cite{p1-first} -- where the agents' next target is the neighbour vertex with the highest idleness, 2) Partitioning strategy (PART) \cite{partition2} -- where the graph is evenly partitioned for each agent to patrol, 3) State exchange Bayesian strategy (SEBS) \cite{dsf1}, a state-of-the-art solution where agents share their locations and next step intention using Bayesian inference to choose their next target \cite{dsf3}, 4) a MARL-based strategy where agents only share their location information (RL-NO-MSG) \cite{rl_recent_1}, and 5) a MARL-based strategy where agents share their location and their next step intention (RL-INTENTION) \cite{p6-RL, rl_recent_3}.

It is worth noting that CR, PART and SEBS do not have battery charging strategies. Therefore, we define that the agents will follow the shortest path to the nearest battery charging station, and the agents will consider the battery remaining similar to the RL-based strategy. For the PART strategy, when an agent is recharging, its assigned partition will be temporarily assigned to nearby agents and reassigned to the supplementary agent once the battery hot-swap is done. The hot-swapped agent will follow the shortest path to its partition. In addition, when an agent is affected by environmental dynamics, e.g., it is moved outside its assigned partition, it will follow the shortest path back to its partition.

We configure the parameters of the environment dynamics based on the settings of real-world experiments described, for example, in  \cite{rl_recent_2}. The specifications of our agents, including battery lifespan, battery hot-swap time, and agents' travelling speed, are based on real-world UAVs (DJI Matrice 300 RTK \cite{dji}). The Matrice 300 RTK has a maximum flight time of $55$ minutes under ideal conditions. Regarding the time for battery hot-swapping, according to its manual, the UAV's batteries must be warmed up before installation. The battery heating process, using a DJI heater \cite{dji-bh}, takes $8$ to $13$ minutes and consumes $3\%$ to $7\%$ of the battery charge. 

We assume that the size of a grid on the graph is $50m$ by $50m$, and the time interval between two discrete timesteps is set to $0.1$ minutes. In ideal conditions, it takes the UAV 1 discrete timestep to patrol to a neighbour vertex (with an $\approx8.33m/s$ flying speed).

The hyperparameters of the agents and environments are shown in Table ~\ref{tab:param-env}.

\renewcommand{\arraystretch}{1.2}
\begin{table}[htb]
    \centering
    \begin{tabular}{ll} 
    \hline
    \textbf{Parameter} & \textbf{Value} \\
    \hline
    $b_{max}$        & $550$ time steps \\
    \hline
    $b^i_{init}$     & $b^i_{init} \sim [90\%, 100\%]$ \\
    \hline
    $b^i_{swap}$     & $b^i_{swap} \sim [80, 150]$ time steps \\
    \hline
    $p^i_{dyn}$      & $p^i_{dyn} \sim [0, 0.05]$ per step and per agent\\
    \hline
    \end{tabular}
    \caption{The values of parameters of the environment and agents.}
    \label{tab:param-env}
\end{table}

\subsection{Model Training}
Trajectories from $8$ parallelly running episodes are collected to train agents. The curriculum learning \cite{curl} method is used, where the patrolling problem gradually increases in complexity as training progresses. Specifically, we increase the maximum number of patrolling agents gradually during training, and hence, the cooperation problems become more complex. This method improves the convergence of the policy \cite{curl} by allowing agents to develop a communication and cooperation strategy gradually for hard patrolling problems. 

The selected values for the reward function parameters are shown in Table ~\ref{tab:rw-param}. The architecture of the agents' actor and critic network is shown in Table ~\ref{tab:Actor} and Table ~\ref{tab:Critic} \footnote{RL-NO-MSG and RL-INTENTION do not have the $Actor_{Comm}$ network.}. Regarding agents' communication, we set agents' message space to be a set from 1 to 16.

The hyperparameters of the MAPPO algorithm are shown in Table ~\ref{tab:mappo}. The models are trained for 1500 episodes, and each episode has a horizon of 5000 steps. For a fair comparison, RL-NO-MSG and RL-INTENTION are trained with the same hyperparameter configuration and with curriculum learning.

\renewcommand{\arraystretch}{1.2}
\begin{table}[htb]
    \centering
    \begin{tabular}{ll} 
    \hline
    \textbf{Parameter} & \textbf{Value} \\
    \hline
    $c_{norm}$       & $200$ \\
    \hline
    $c_{r'_p}$       & $0.5$ \\
    \hline
    $c_{d}$          & $\frac{50}{\text{num of maximum patrolling agents}}$\\
    \hline
    $c_{p_b}$        & $50$ \\
    \hline
    $m$              & $20$ \\
    \hline
    $c_{r_{b_2}}$    & \makecell[l]{$1$ for MAP A, C, D \\ $0.5$ for MAP B} \\
    \hline
    $c_{p_c}$        & $1$\\
    \hline
    
    \end{tabular}
    \caption{The parameters of the reward functions.}
    \label{tab:rw-param}
\end{table}

\renewcommand{\arraystretch}{1.2}
\begin{table}[htb]
    \centering
    \begin{tabular}{ccc} 
    \hline
    \textbf{Layer} & \textbf{Parameter} & \textbf{Activation Function} \\
    \hline
    Conv &  \makecell{ic=2, oc=4, ks=(3,3), \\ s=1, p=0} & Tanh \\
    \hline
    Conv &  \makecell{ic=4, oc=8, ks=(3,3), \\ s=1, p=0} & Tanh \\
    \hline
    Dense & 519 $\times$ 512 & Tanh \\
    \hline
    Dense & 512 $\times$ 341 & Tanh \\
    \hline
    Dense & 341 $\times$ 227 & Tanh \\
    \hline
    Dense & \makecell{227 $\times$ 1 - $Actor_{act}$ \\ 227 $\times$ 16 - $Actor_{Comm}$} & None \\
    \hline
    \end{tabular}
    \caption{Neural network architecture of the agent's actor network ($Actor_{act}$ and $Actor_{Comm}$). "Conv" -- convolutional layer, "Dense" -- dense layer, "ic" -- "input channel". "oc" -- "output channel". "ks" -- "kernel size". "s" -- "stride". "p" -- "padding". }
    \label{tab:Actor}
\end{table}

\renewcommand{\arraystretch}{1.2}
\begin{table}[htb]
    \centering
    \begin{tabular}{ccc}     
    \hline
    \textbf{Layer} & \textbf{Parameter} & \textbf{Activation Function} \\
    \hline
    Conv &  \makecell{ic=2, oc=4, ks=(3,3), \\ s=1, p=0} & Tanh \\
    \hline
    Conv &  \makecell{ic=4, oc=8, ks=(3,3), \\ s=1, p=0} & Tanh \\
    \hline
    Dense & 527 $\times$ 512 & Tanh \\
    \hline
    Dense & 512 $\times$ 341 & Tanh \\
    \hline
    Dense & 341 $\times$ 227 & Tanh \\
    \hline
    Dense & 227 $\times$ 1 & None \\
    \hline
    \end{tabular}
    \caption{Neural network architecture of the agent's critic network.}
    \label{tab:Critic}
\end{table}

\setlength{\tabcolsep}{5pt}
\begin{table}[htb]
    \centering
    \begin{tabular}{cc}
        \hline
        \textbf{Parameter} & \textbf{Parameter Value} \\
        \hline
        $\gamma$           & 0.95 \\
        \hline
        GAE $\lambda$      & 0.95 \\
        \hline
        Policy clip        & 0.15 \\
        \hline
        Number of Batches  & 50 \\
        \hline
        Epoch              & 3 \\
        \hline
        Entropy coefficient & 0.002  \\ 
        \hline
        Learning rate & \makecell[l]{2e-4 before 1000 episodes \\ and 1e-4 afterwards.}  \\
        \hline
        \makecell[l]{Number of patrolling  \\ agents in 8 parallel \\ episodes. "ep" is the \\ number of training \\ episodes} & \makecell[l]{ 1, 1, 1, 1, 2, 2, 2, 2 $\mid$ ep $<$ 200 \\  1, 1, 1, 1, 2, 2, 3, 3 $\mid$ 200 $\leq$ ep $<$ 400 \\ 1, 1, 1, 1, 2, 3, 3, 4 $\mid$ 400 $\leq$ ep $<$ 600 \\ 1, 1, 1, 1, 2, 3, 4, 5 $\mid$ 600 $\leq$ ep} \\
        \hline
        \end{tabular}
    \caption{Values of MAPPO hyperparameters.}
    \label{tab:mappo}
\end{table}

\begin{figure}[!hbt]
    \begin{subfigure}[b]{.48\linewidth}
        \centering
        \includegraphics[width=.95\linewidth]{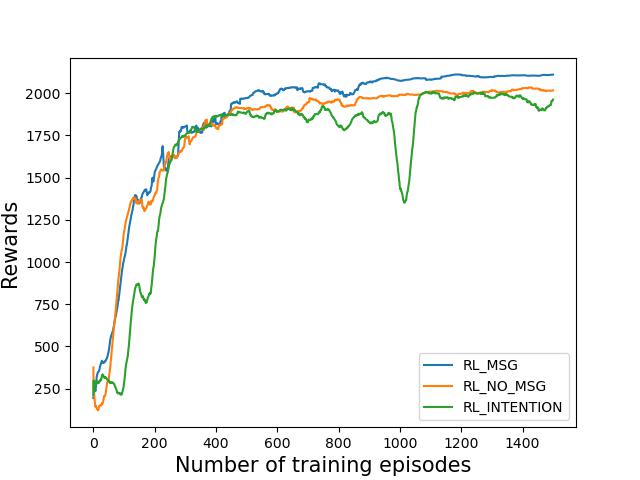}
        \caption{MAP A}
    \end{subfigure}\hfill
    \begin{subfigure}[b]{.48\linewidth}
        \centering
        \includegraphics[width=.95\linewidth]{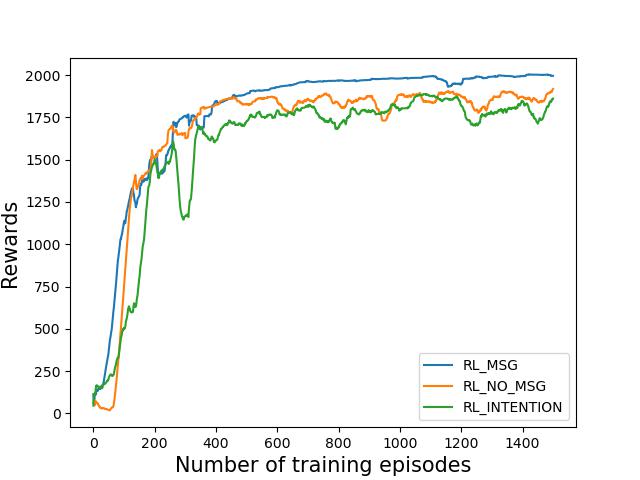}
        \caption{MAP B}
    \end{subfigure}
    
    \begin{subfigure}[b]{.48\linewidth}
        \centering
        \includegraphics[width=.95\linewidth]{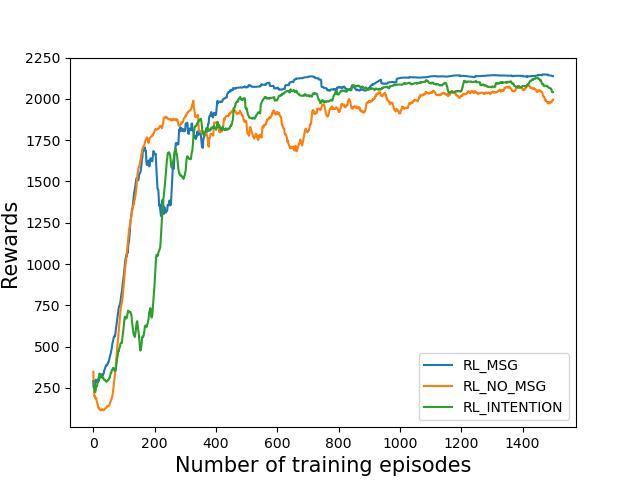}
        \caption{MAP C}
    \end{subfigure}\hfill
    \begin{subfigure}[b]{.48\linewidth}
        \centering
        \includegraphics[width=.95\linewidth]{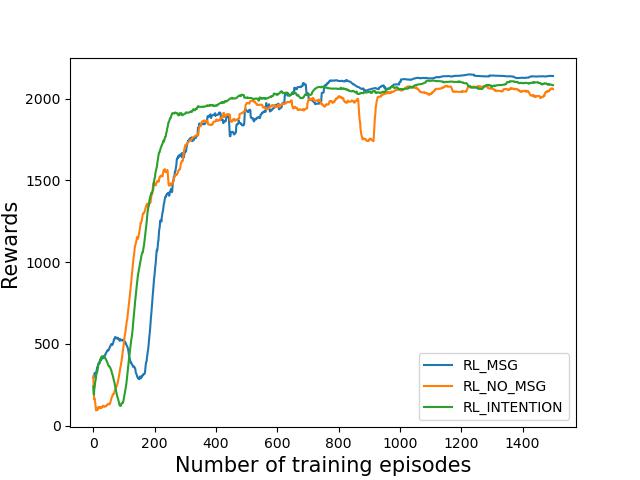}
        \caption{MAP D}
    \end{subfigure}%
    \caption{Four patrolling maps. The numbers represent the priorities of vertices, while the un-numbered vertices have priority 0.}
     \label{fig:training}
\end{figure}

Fig.~\ref{fig:training} demonstrates the agents' cumulative reward for RL-MSG, RL-NO-MSG, and RL-INTENTION methods. The solid curves show the average of agents' last $50$ cumulative reward. The results indicate that the proposed MARL-based strategy can achieve a higher cumulative reward compared with strategies where agents are not allowed to communicate or where agents can only share their intentions. 

\subsection{Battery Recharging Performance Evaluation} \label{Sec:BCP}

A model's battery recharging performance is evaluated based on the agents' battery failure rate and their amount of battery remaining when recharging. The data is collected from 10 tests, each containing 100 test episodes. Each episode has a horizon of $14,400$ steps simulating approximately a real-world 24-hour patrolling scenario, i.e., each test calculates the average battery failure rate over 100 days.

For the CR, PART, and SEBS strategies, we define a battery charging strategy with a near-zero \footnote{The environmental dynamics may result in agents failing to reach the battery charging station} failure rate. Agents dynamically plan their shortest path to the battery charging station and avoid deadlock situations, where no agents can move as their intended directions are blocked by other agents.

Table~\ref{tab:bf} and Table~\ref{tab:blv} show the amount of agents' battery remaining when recharging and their battery failure rate on four maps patrolling with $1$ to $5$ agents. From Table~\ref{tab:blv}, we observe that the result indicates that all MARL-based strategies have learned to recharge with battery charge remaining as required. Table~\ref{tab:bf} shows that RL-MSG has the lowest average battery failure rate across all reinforcement learning methods with a near-zero failure rate on Maps A, C and D. 

On the other hand, RL-INTENTION has no better battery charging performance compared with RL-NO-MSG. This indicates that pre-defined communication strategies may not result in optimal patrolling strategies in complex cooperation scenarios. From the results, we can conclude that the proposed agents have successfully developed a strategy that coordinates how multiple agents can recharge.




\subsection{Patrolling Performance Evaluation} \label{Sec:PP}

A model's patrolling performance is evaluated based on $AVG^h(G)$, $\overline{MAX^h(G)}$ averaged over 100 successfully completed test episodes without battery failure. Each episode has a horizon of $14,400$ steps.

Table ~\ref{tab:max} and Table ~\ref{tab:mean} show the results for the patrolling performance. On average, RL-MSG has the most optimal $\overline{MAX^h(G)}$ on all maps with different numbers of patrolling agents and has a $AVG^h(G)$ no worse than other strategies.  

It is worth noting that RL-MSG outperforms RL-NO-MSG and RL-INTENTION in single-agent scenarios when measuring  $\overline{MAX^h(G)}$ since the communication has no contribution to patrolling. This indicates that RL-NO-MSG and RL-INTENTION agents are less greedy when visiting vertices with high priority/idleness. When patrolling with multiple uncoordinated agents, if all agents are greedy in visiting vertices with high priority/idleness, this could lead to other vertices not being patrolled often and result in a poor overall patrolling performance. Therefore, RL-NO-MSG and RL-INTENTION may choose a sub-optimal strategy where agents are less greedy in visiting high-prioritised vertices, assuring each node is visited with an even frequency. 

On the other hand, as the RL-MSG has a low $\overline{MAX^h(G)}$, this indicates that agents have developed a communication strategy that allows them to negotiate and evenly assign each other patrolling targets, assuring high prioritised vertices are visited frequently without leaving low prioritised vertices unvisited. Therefore, from this result, we can conclude that the proposed agents have the ability to communicate and cooperate, resulting in more optimal patrolling strategies.

In addition, the RL-INTENTION agents perform even worse than RL-MSG, indicating that manually defined communication information may not always have a positive effect in finding optimal patrolling strategies.

\subsection{Collision/Congestion Avoidance Performance Evaluation} \label{Sec:collision}

A model's collision/congestion avoidance performance is evaluated based on the number of collisions that occur among agents. The data is averaged over 100 successfully completed test episodes without battery failure. Each episode has a horizon of $14,400$ steps.

Table \ref{tab:collision} shows the number of agents' collisions in a simulated 24-hour patrolling scenario using different policies. The CR strategy has no coordination between agents and hence has the highest collision rate. In PART strategies, agents are patrolling non-overlapping areas, so the collisions mostly occur when agents are on their way to recharge, indicating congestion situations. For example, some agents try to enter the battery charging station while other agents are trying to leave. The result shows that PART strategy collision/congestion avoidance performance is map-dependent, since agents have the best collision avoidance performance on MAP B, but worse performance than RL-MSG strategies on MAP A and C. The agents also frequently collide on MAP D. 

Regarding the RL-based strategy collision/congestion avoidance performance, RL-MSG has the least number of collisions among all strategies on MAP A, C and D and has the best performance among MARL-based strategies on MAP B. Therefore, we can conclude that the proposed agents have developed a communication protocol to effectively avoid collisions and congestion.

In addition, the results show that sharing agent intentions will not improve the agents' ability to avoid collisions. Indeed sometimes, it can cause agents to perform even worse.

\subsection{Fault tolerance Evaluation}

According to the results in Section \ref{Sec:BCP}, Section \ref{Sec:collision}, and Section \ref{Sec:PP}, on average, the RL-MSG strategy has the optimal performance on all maps based on 1 to 5  patrolling agents. It indicates that the cooperation strategy developed by agents can work with different numbers of agents, and the patrolling system can keep functioning when one or multiple agents fail. Therefore, the fault tolerance of the proposed method is demonstrated.


\begin{table*}[htbp]
        \sisetup{round-mode=places, round-precision=5}
        \centering
        \resizebox{0.9\textwidth}{!}{%
            \begin{tabular}{lllllll}%
            MAP A & & & & & & \\
            \sisetup{round-mode=places, round-precision=5}
            \bfseries $n$ & \bfseries $CR$ & \bfseries $PART$ & \bfseries $SEBS$ & \bfseries $RL-INTENTION$ & \bfseries $RL-NO-MSG$ & \bfseries $RL-MSG$
            \csvreader[head to column names]{A-BF}{}
            {\\\hline \NA & \CR & \PART & \SEBS & \INTENTION & \NOMSG & \MSG}
            \end{tabular}
        }
        \resizebox{0.9\textwidth}{!}{%
            \begin{tabular}{lllllll}%
            MAP B & & & & & & \\
            \sisetup{round-mode=places, round-precision=5}
            \bfseries $n$ & \bfseries $CR$ & \bfseries $PART$ & \bfseries $SEBS$ & \bfseries $RL-INTENTION$ & \bfseries $RL-NO-MSG$ & \bfseries $RL-MSG$
            \csvreader[head to column names]{B-BF}{}
            {\\\hline \NA & \CR & \PART & \SEBS & \INTENTION & \NOMSG & \MSG}
            \end{tabular}
        }
        \resizebox{0.9\textwidth}{!}{%
            \begin{tabular}{lllllll}%
            MAP C & & & & & & \\
            \sisetup{round-mode=places, round-precision=5}
            \bfseries $n$ & \bfseries $CR$ & \bfseries $PART$ & \bfseries $SEBS$ & \bfseries $RL-INTENTION$ & \bfseries $RL-NO-MSG$ & \bfseries $RL-MSG$
            \csvreader[head to column names]{C-BF}{}
            {\\\hline \NA & \CR & \PART & \SEBS & \INTENTION & \NOMSG & \MSG}
            \end{tabular}
        }
        \resizebox{0.9\textwidth}{!}{%
            \begin{tabular}{lllllll}%
            MAP D & & & & & & \\
            \sisetup{round-mode=places, round-precision=5}
            \bfseries $n$ & \bfseries $CR$ & \bfseries $PART$ & \bfseries $SEBS$ & \bfseries $RL-INTENTION$ & \bfseries $RL-NO-MSG$ & \bfseries $RL-MSG$
            \csvreader[head to column names]{D-BF}{}
            {\\\hline \NA & \CR & \PART & \SEBS & \INTENTION & \NOMSG & \MSG}
            \end{tabular}
        }
        \caption{Agents' battery failure rate. Run with $1$ to $5$ number of patrolling agents on four maps}
        \label{tab:bf}
\end{table*}

\begin{table*}[htbp]
        \sisetup{round-mode=places, round-precision=5}
        \centering
        \resizebox{0.9\textwidth}{!}{%
            \begin{tabular}{lllllll}%
            MAP A & & & & & & \\
            \sisetup{round-mode=places, round-precision=5}
            \bfseries $n$ & \bfseries $CR$ & \bfseries $PART$ & \bfseries $SEBS$ & \bfseries $RL-INTENTION$ & \bfseries $RL-NO-MSG$ & \bfseries $RL-MSG$
            \csvreader[head to column names]{A-BLv}{}
            {\\\hline \NA & \CR & \PART & \SEBS & \INTENTION & \NOMSG & \MSG}
            \end{tabular}
        }
        \resizebox{0.9\textwidth}{!}{%
            \begin{tabular}{lllllll}%
            MAP B & & & & & & \\
            \sisetup{round-mode=places, round-precision=5}
            \bfseries $n$ & \bfseries $CR$ & \bfseries $PART$ & \bfseries $SEBS$ & \bfseries $RL-INTENTION$ & \bfseries $RL-NO-MSG$ & \bfseries $RL-MSG$
            \csvreader[head to column names]{B-BLv}{}
            {\\\hline \NA & \CR & \PART & \SEBS & \INTENTION & \NOMSG & \MSG}
            \end{tabular}
        }
        \resizebox{0.9\textwidth}{!}{%
            \begin{tabular}{lllllll}%
            MAP C & & & & & & \\
            \sisetup{round-mode=places, round-precision=5}
            \bfseries $n$ & \bfseries $CR$ & \bfseries $PART$ & \bfseries $SEBS$ & \bfseries $RL-INTENTION$ & \bfseries $RL-NO-MSG$ & \bfseries $RL-MSG$
            \csvreader[head to column names]{C-BLv}{}
            {\\\hline \NA & \CR & \PART & \SEBS & \INTENTION & \NOMSG & \MSG}
            \end{tabular}
        }
        \resizebox{0.9\textwidth}{!}{%
            \begin{tabular}{lllllll}%
            MAP D & & & & & & \\
            \sisetup{round-mode=places, round-precision=5}
            \bfseries $n$ & \bfseries $CR$ & \bfseries $PART$ & \bfseries $SEBS$ & \bfseries $RL-INTENTION$ & \bfseries $RL-NO-MSG$ & \bfseries $RL-MSG$
            \csvreader[head to column names]{D-BLv}{}
            {\\\hline \NA & \CR & \PART & \SEBS & \INTENTION & \NOMSG & \MSG}
            \end{tabular}
        }
        \caption{Agents' battery remaining when recharging. Run with $1$ to $5$ number of patrolling agents on four maps}
        \label{tab:blv}
\end{table*}


\begin{table*}[htbp]
        \sisetup{round-mode=places, round-precision=5}
        \centering
        \resizebox{0.9\textwidth}{!}{%
            \begin{tabular}{lllllll}%
            MAP A & & & & & & \\
            \sisetup{round-mode=places, round-precision=5}
            \bfseries $n$ & \bfseries $CR$ & \bfseries $PART$ & \bfseries $SEBS$ & \bfseries $RL-INTENTION$ & \bfseries $RL-NO-MSG$ & \bfseries $RL-MSG$
            \csvreader[head to column names]{A-MAX}{}
            {\\\hline \NA & \CR & \PART & \SEBS & \INTENTION & \NOMSG & \MSG}
            \end{tabular}
        }
        \resizebox{0.9\textwidth}{!}{%
            \begin{tabular}{lllllll}%
            MAP B & & & & & & \\
            \sisetup{round-mode=places, round-precision=5}
            \bfseries $n$ & \bfseries $CR$ & \bfseries $PART$ & \bfseries $SEBS$ & \bfseries $RL-INTENTION$ & \bfseries $RL-NO-MSG$ & \bfseries $RL-MSG$
            \csvreader[head to column names]{B-MAX}{}
            {\\\hline \NA & \CR & \PART & \SEBS & \INTENTION & \NOMSG & \MSG}
            \end{tabular}
        }
        \resizebox{0.9\textwidth}{!}{%
            \begin{tabular}{lllllll}%
            MAP C & & & & & & \\
            \sisetup{round-mode=places, round-precision=5}
            \bfseries $n$ & \bfseries $CR$ & \bfseries $PART$ & \bfseries $SEBS$ & \bfseries $RL-INTENTION$ & \bfseries $RL-NO-MSG$ & \bfseries $RL-MSG$
            \csvreader[head to column names]{C-MAX}{}
            {\\\hline \NA & \CR & \PART & \SEBS & \INTENTION & \NOMSG & \MSG}
            \end{tabular}
        }
        \resizebox{0.9\textwidth}{!}{%
            \begin{tabular}{lllllll}%
            MAP D & & & & & & \\
            \sisetup{round-mode=places, round-precision=5}
            \bfseries $n$ & \bfseries $CR$ & \bfseries $PART$ & \bfseries $SEBS$ & \bfseries $RL-INTENTION$ & \bfseries $RL-NO-MSG$ & \bfseries $RL-MSG$
            \csvreader[head to column names]{D-MAX}{}
            {\\\hline \NA & \CR & \PART & \SEBS & \INTENTION & \NOMSG & \MSG}
            \end{tabular}
        }
        \caption{Agents' patrolling performance: $\overline{MAX^h(G)}$. Run with $1$ to $5$ number of patrolling agents on four maps}
        \label{tab:max}
\end{table*}

\begin{table*}[htbp]
        \sisetup{round-mode=places, round-precision=5}
        \centering
        \resizebox{0.9\textwidth}{!}{%
            \begin{tabular}{lllllll}%
            MAP A & & & & & & \\
            \sisetup{round-mode=places, round-precision=5}
            \bfseries $n$ & \bfseries $CR$ & \bfseries $PART$ & \bfseries $SEBS$ & \bfseries $RL-INTENTION$ & \bfseries $RL-NO-MSG$ & \bfseries $RL-MSG$
            \csvreader[head to column names]{A-MEAN}{}
            {\\\hline \NA & \CR & \PART & \SEBS & \INTENTION & \NOMSG & \MSG}
            \end{tabular}
        }
        \resizebox{0.9\textwidth}{!}{%
            \begin{tabular}{lllllll}%
            MAP B & & & & & & \\
            \sisetup{round-mode=places, round-precision=5}
            \bfseries $n$ & \bfseries $CR$ & \bfseries $PART$ & \bfseries $SEBS$ & \bfseries $RL-INTENTION$ & \bfseries $RL-NO-MSG$ & \bfseries $RL-MSG$
            \csvreader[head to column names]{B-MEAN}{}
            {\\\hline \NA & \CR & \PART & \SEBS & \INTENTION & \NOMSG & \MSG}
            \end{tabular}
        }
        \resizebox{0.9\textwidth}{!}{%
            \begin{tabular}{lllllll}%
            MAP C & & & & & & \\
            \sisetup{round-mode=places, round-precision=5}
            \bfseries $n$ & \bfseries $CR$ & \bfseries $PART$ & \bfseries $SEBS$ & \bfseries $RL-INTENTION$ & \bfseries $RL-NO-MSG$ & \bfseries $RL-MSG$
            \csvreader[head to column names]{C-MEAN}{}
            {\\\hline \NA & \CR & \PART & \SEBS & \INTENTION & \NOMSG & \MSG}
            \end{tabular}
        }
        \resizebox{0.9\textwidth}{!}{%
            \begin{tabular}{lllllll}%
            MAP D & & & & & & \\
            \sisetup{round-mode=places, round-precision=5}
            \bfseries $n$ & \bfseries $CR$ & \bfseries $PART$ & \bfseries $SEBS$ & \bfseries $RL-INTENTION$ & \bfseries $RL-NO-MSG$ & \bfseries $RL-MSG$
            \csvreader[head to column names]{D-MEAN}{}
            {\\\hline \NA & \CR & \PART & \SEBS & \INTENTION & \NOMSG & \MSG}
            \end{tabular}
        }
        \caption{Agents' patrolling performance: $AVG^h(G)$. Run with $1$ to $5$ number of patrolling agents on four maps}
        \label{tab:mean}
\end{table*}


\begin{table*}[htbp]
        \sisetup{round-mode=places, round-precision=5}
        \centering
        \resizebox{0.9\textwidth}{!}{%
            \begin{tabular}{lllllll}%
            MAP A & & & & & & \\
            \sisetup{round-mode=places, round-precision=5}
            \bfseries $n$ & \bfseries $CR$ & \bfseries $PART$ & \bfseries $SEBS$ & \bfseries $RL-INTENTION$ & \bfseries $RL-NO-MSG$ & \bfseries $RL-MSG$
            \csvreader[head to column names]{A-COLLISION}{}
            {\\\hline \NA & \CR & \PART & \SEBS & \INTENTION & \NOMSG & \MSG}
            \end{tabular}
        }
        \resizebox{0.9\textwidth}{!}{%
            \begin{tabular}{lllllll}%
            MAP B & & & & & & \\
            \sisetup{round-mode=places, round-precision=5}
            \bfseries $n$ & \bfseries $CR$ & \bfseries $PART$ & \bfseries $SEBS$ & \bfseries $RL-INTENTION$ & \bfseries $RL-NO-MSG$ & \bfseries $RL-MSG$
            \csvreader[head to column names]{B-COLLISION}{}
            {\\\hline \NA & \CR & \PART & \SEBS & \INTENTION & \NOMSG & \MSG}
            \end{tabular}
        }
        \resizebox{0.9\textwidth}{!}{%
            \begin{tabular}{lllllll}%
            MAP C & & & & & & \\
            \sisetup{round-mode=places, round-precision=5}
            \bfseries $n$ & \bfseries $CR$ & \bfseries $PART$ & \bfseries $SEBS$ & \bfseries $RL-INTENTION$ & \bfseries $RL-NO-MSG$ & \bfseries $RL-MSG$
            \csvreader[head to column names]{C-COLLISION}{}
            {\\\hline \NA & \CR & \PART & \SEBS & \INTENTION & \NOMSG & \MSG}
            \end{tabular}
        }
        \resizebox{0.9\textwidth}{!}{%
            \begin{tabular}{lllllll}%
            MAP D & & & & & & \\
            \sisetup{round-mode=places, round-precision=5}
            \bfseries $n$ & \bfseries $CR$ & \bfseries $PART$ & \bfseries $SEBS$ & \bfseries $RL-INTENTION$ & \bfseries $RL-NO-MSG$ & \bfseries $RL-MSG$
            \csvreader[head to column names]{D-COLLISION}{}
            {\\\hline \NA & \CR & \PART & \SEBS & \INTENTION & \NOMSG & \MSG}
            \end{tabular}
        }
        \caption{Agents' collision avoidance performance. Run with $1$ to $5$ number of patrolling agents on four maps}
        \label{tab:collision}
\end{table*}


\clearpage
\section{Conclusion}
This work has proposed a deep multi-agent reinforcement learning-based patrolling approach that allows agents to develop their own communication protocol to cooperate during multi-agent patrolling problems. Agents can communicate to avoid collisions, assign different patrolling targets to each other, and negotiate battery recharging schemes to support continuous patrolling. The patrolling system can remain functioning when one or multiple agents fail. We propose a modified state-of-the-art reinforcement learning algorithm (Proximal Policy Optimization) to train agents by adopting a reinforced inter-agent learning (RIAL) method \cite{RIAL}. Simulation experiments show that our approach has improved patrolling performance and collision avoidance performance compared with other state-of-the-art solutions based on a Conscientious Reactive strategy, Partitioning strategy, State exchange Bayesian strategy, and RL-based approach where agents only share their location information and their next step intention. In addition, the proposed solution has the lowest battery failure rate among the RL-based solutions.

There are several areas that could be explored in future extensions to the work, including:
\begin{itemize}
    \item patrolling with agents with different specifications and policies; 
    \item supporting patrolling policies that allow agents to patrol different maps;
    \item deploying the solution to real-world patrolling vehicle scenarios.
\end{itemize}

\clearpage
\section{Acknowledgements}

This study was supported by a \emph{Melbourne Research Scholarship} from the University of Melbourne, VIC, Australia.\\

\clearpage
\bibliographystyle{plain} 
\bibliography{reference}

\end{document}